\documentclass{article}

\usepackage{arxiv}

\usepackage[utf8]{inputenc} % allow utf-8 input
\usepackage[T1]{fontenc}    % use 8-bit T1 fonts
\usepackage{hyperref}       % hyperlinks
\usepackage{url}            % simple URL typesetting
\usepackage{booktabs}       % professional-quality tables
\usepackage{amsfonts}       % blackboard math symbols
\usepackage{nicefrac}       % compact symbols for 1/2, etc.
\usepackage{microtype}      % microtypography
\usepackage{lipsum}		% Can be removed after putting your text content
\usepackage{graphicx}
\usepackage{doi}

\usepackage{enumerate,enumitem}
\usepackage{subfigure}
\usepackage{color,xcolor}
\usepackage{multicol,multirow}
\usepackage{colortbl}
\usepackage{amsmath}
\usepackage{booktabs}
\usepackage{algorithm}
\usepackage{algorithmic}

\title{Conformal Uncertainty Indicator for Continual Test-Time Adaptation}

\author{
Fan Lyu$^1$, Hanyu Zhao$^2$, Ziqi Shi$^{3}$,Ye Liu$^4$,Fuyuan Hu$^5$,Zhang Zhang$^{1*}$, Liang Wang$^1$\\\\
$^1$New Laboratory of Pattern Recognition, Institute of Automation, Chinese Academy of Sciences\\
$^2$China Nuclear Power Engineering Co., LTD\\
$^3$College of Intelligence and Computing, Tianjin University \\
$^4$School of Computer Science \& Engineering, LinYi University \\
$^5$School of Electric \& Information Engineering, Suzhou University of Science and Technology\\
\texttt{fan.lyu@cripac.ia.ac.cn, zzhang@nlpr.ia.ac.cn}
}

\hypersetup{
pdftitle={A template for the arxiv style},
pdfsubject={q-bio.NC, q-bio.QM},
pdfauthor={David S.~Hippocampus, Elias D.~Striatum},
pdfkeywords={First keyword, Second keyword, More},
}

\def\ie{\textit{i.e.}}

\def\mbb{\mathbb}
\def\mc{\mathcal}

\begin{document}
\maketitle

\begin{abstract}
Continual Test-Time Adaptation (CTTA) aims to adapt models to sequentially changing domains during testing, relying on pseudo-labels for self-adaptation. However, incorrect pseudo-labels can accumulate, leading to performance degradation. To address this, we propose a Conformal Uncertainty Indicator (CUI) for CTTA, leveraging Conformal Prediction (CP) to generate prediction sets that include the true label with a specified coverage probability.
Since domain shifts can lower the coverage than expected, making CP unreliable, we dynamically compensate for the coverage by measuring both domain and data differences. 
Reliable pseudo-labels from CP are then selectively utilized to enhance adaptation. Experiments confirm that CUI effectively estimates uncertainty and improves adaptation performance across various existing CTTA methods.
\end{abstract}

% keywords can be removed
% \keywords{First keyword \and Second keyword \and More}

\section{Introduction}

Recently, Continual Test-Time Adaptation (CTTA)~\cite{wang2022continual} has garnered significant attention for enabling trained models to handle various unknown test domain shifts through self-adaptation. 
This innovative approach aims to enhance model robustness and adaptability during the testing phase, addressing the dynamic nature of real-world data, such as autonomous driving~\cite{sojka2023ar} and medical imagining~\cite{chen2024each}.
However, a critical challenge arises in many testing scenarios where \textit{the cost of incorrect predictions is prohibitively high}, such as autonomous driving and medical scenarios. 
Unreliable predictions in self-adaptation may lead to severe error accumulation, decreasing the model’s performance. 
Therefore, effectively measuring the uncertainty of model outputs becomes crucial to mitigate losses and allow for human intervention or termination.

\begin{figure}[t]
    \centering
    \includegraphics[width=.75\linewidth]{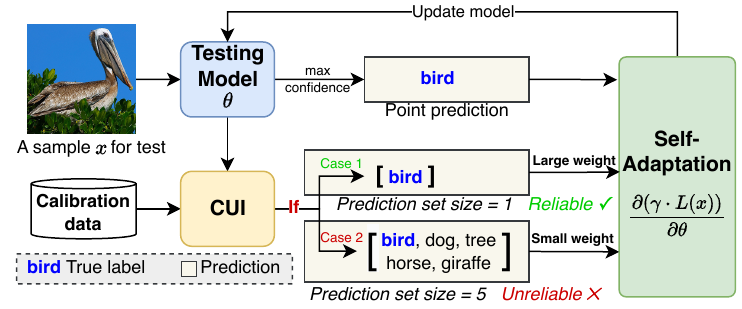}
    \caption{In the task of CTTA, a test sample $x$ may be drawn from a different distribution in a long-term testing phase.
    Traditional methods rely on the self-adaptation based on the prediction and ignore the uncertainty may cause error accumulation.
    CUI provides a technique of uncertainty measurement based on CP.
    For the test sample, if CUI outputs a prediction set with small sizes ($>0$), it is regarded as reliable and yields a large loss weight in adaptation.
    Large prediction sets mean unreliable prediction.
    The coverage means that the true label is included in the prediction set.
    The example image is sampled from ImageNet~\cite{deng2009imagenet}.
    }
    \vspace{-10px}
    \label{fig:fig1}
\end{figure}

Some uncertainty estimation methods are based on Bayes rule, such as Bayes approximation~\cite{maddox2019simple} and Monte Carlo dropout~\cite{gal2016dropout}, requiring high computational complexity or rely on model selection, thus difficult to be applied to testing time.
Moreover, some methods directly use the output logits to form uncertainties such as entropy~\cite{shi2024controllable}, which may suffer from confidence dilemma that unreliable logits give unreliable uncertainty estimations.
In contrast, Conformal Prediction (CP)~\cite{vovk2005algorithmic} offers a promising solution for measuring uncertainty in predictions, which produces set-valued predictions that serve as a wrapper around existing models.
CP has the following compelling advantages.
First, CP is \textit{model-agnostic}, which means it does not require any assumptions about the model, making it applicable to any pre-trained model without necessitating modifications.
Second, CP yields \textit{controllable coverage}, which means CP allows the true label coverage probability to be pre-specified and ensures that this probability is met.
These advantages meet the scenario of CTTA that continuously measures the output uncertainties for a pre-trained model in testing time.

However, incorporating CP into unsupervised CTTA presents significant challenges. 
Traditional CP requires the assumption of data exchangeability, which refers to the assumption that the order in which the data points are observed does not matter.
The assumption is violated under domain shift conditions, thus leading to the coverage gap issue~\cite{barber2023conformal}.
The coverage gap means that the uncertainty estimation is under the coverage much less than the given expectation. 
That is, \textit{the uncertainty estimation is not trustworthy in this situation}.

In this paper, we explore the feasibility of using CP in testing scenarios by addressing the coverage gap challenge and propose a simple \textit{plug-an-play} uncertainty measurement method named Conformal Uncertainty Indicator (CUI).
\textit{The goal of CUI is to output the uncertainty of testing for each test example with a given trained model}.
\textit{The key motivation for the design of CUI is to compensate the coverage gap when domain shifts and output reliable uncertainty level}.
CUI leverages a static source calibration set with labels from pre-training. 
During testing, it evaluates uncertainty by measuring domain shifts from model and data differences. A test sample's quantile is computed based on the calibration set, adjusted by the domain shift to enhance coverage. Finally, a non-conformity threshold is derived from the adjusted quantile to generate a prediction set, with its size reflecting the uncertainty level.
Moreover, based on the CP results, we design a simple enhanced adaptation method on confident test samples, which can be applied to existing CTTA methods.
We find applying more adaptations on samples with reliable predictions will get good testing performance.
{As shown in Fig.~\ref{fig:fig1}, a traditional CTTA block consists of a point prediction and an adaptation, the proposed CUI provides the testing uncertainty and helps the adaptation.}
We evaluate on three benchmark datasets and find that the proposed CUI can better evaluate the test uncertainty than other CP methods.
By integrating the CP-based adaptation strategy, existing methods achieve better reliability and robustness of model predictions in dynamic and uncertain test environments.
Our contributions are three-fold:
\begin{enumerate}[label=(\arabic*),left=0pt,itemsep=0pt]
    \vspace{-5px}
    \item We propose a simple uncertainty estimation method CUI for CTTA to measure the test uncertainty for each test prediction. CUI is model-agnostic and relies only on a small size of calibration set.
    \item We propose an adaptation method based on the CUI estimation, which enhances the reliable test adaptation.
    \item We evaluate our method on benchmark datasets and help multiple existing CTTA methods measure their test uncertainty and achieve better performance via our adaptation strategy.
\end{enumerate}

\section{Related Work}

\textbf{Continual Test-Time Adaptation}.
Test-Time Adaptation (TTA) allows source-free, online model adaptation to target domain characteristics \cite{jain2011online,sun2020test,wang2020tent}. CTTA~\cite{wang2022continual,lyu2024variational,tan2024less} extends TTA to continuously changing target domains, addressing long-term adaptation but facing error accumulation challenges~\cite{tarvainen2017mean,wang2022continual}. 
Prolonged reliance on unsupervised loss during long-term adaptation risks error accumulation and forgetting source knowledge, hindering accurate classification of source-like test samples. Most existing methods address these issues by enhancing the source model's confidence during testing.
To address error accumulation, mean-teacher methods~\cite{tarvainen2017mean} align the student with the teacher model, updating the teacher via moving averages. To mitigate forgetting, augmentation-averaged predictions~\cite{wang2022continual,brahma2023probabilistic,dobler2023robust,yang2023exploring} enhance the teacher's confidence against out-of-distribution samples. 
Contrastive loss~\cite{dobler2023robust,chakrabarty2023sata} preserves learned semantics, while some methods prioritize restoring source parameters~\cite{wang2022continual,brahma2023probabilistic}.
Though the above methods keep the model from vague pseudo labels, they may suffer from overly confident predictions that are less calibrated.
To mitigate this issue, it is helpful to estimate the uncertainty in the model.

\noindent
\textbf{Conformal Prediction}.
CP is a robust framework for quantifying uncertainty in machine learning, especially in high-stakes applications where reliability is crucial. 
CP generates prediction sets that contain the true outcome with a specified probability, without relying on assumptions about the underlying data distribution. CP focuses on the concept of exchangeability and the use of nonconformity scores~\cite{vovk2005algorithmic}.
CP has been applied to a wide range of problems, including medical diagnostics~\cite{caruana2015intelligible}, autonomous vehicles~\cite{lekeufack2023conformal}, and financial decision-making, where the quantification of uncertainty is critical for safety and trust.
Researchers have extended conformal prediction to handle more complex scenarios such as risk control \cite{farinhas2023non}.
However, to the best of our knowledge, conformal prediction has not yet been applied to the CTTA task, whereas estimating uncertainty in test results is crucial in long-term testing environments.

\section{Preliminary: CTTA and CP}

\subsection{Continual Test-Time Adaptation (CTTA)}
Given a classification model pre-trained on a source domain, CTTA methods adapt the source model to the unlabeled target data, where the domain continuously changes.
Because the adaptation is conducted during test time, which means the model needs to output the prediction immediately then update the model.
The unsupervised dataset of target domains are denoted as $\mathcal{D}^k = \{x^k_m\}^{N^k}_{m=1}$, where $k$ is the target domain index.
For each test sample, CTTA conducts two major operations: testing and adaptation.
For testing, the model needs to output the prediction of the model.
For adaptation, the model needs to adapt to the testing sample without ground truth.
{In many applications such as autonomous driving and medical diagnosis, the cost of misprediction is high, thus it is crucial to estimate the uncertainty of test results.
In this paper, we use conformal prediction to evaluate prediction uncertainties}.

\subsection{Conformal Prediction and Coverage Gap Issue}

We first introduce CP under a multi-class classification task with total $K$ classes.
Let $\mc{X}$ be the input space and $\mc{Y}:= \{1, \cdots, K\}$ be the label space. 
We use ${\pi}: \mc{X} \rightarrow \mbb{R}^K$ to denote the pre-trained neural network that is used to predict the label of a test sample.
The model prediction in this classification task is generally made as
\begin{equation}
\hat{y}=\arg\max_{y\in\mathcal{Y}}{\pi}({y|x}),
\end{equation}
where ${\pi}({y|x})$ can be seen as the confidence of that $x$ being labeled to class $y$.
Only predicting point labels (only labels with max confidence will be selected) from the model outputs does not yield prediction uncertainty.
To provide an uncertainty guarantee for the model performance, CP~\cite{vovk2005algorithmic} is designed to produce prediction sets containing true labels with a desired probability. 
Standard CP takes a test sample $x\in \mc{X}^{\mathrm{test}}$, creating a prediction set $\mc{P}(x)\subseteq\mc{Y}^{\mathrm{test}} $ and satisfying \textbf{marginal coverage} for the true label $y\in\mc{Y}^{\mathrm{test}}$:
\begin{equation}
    \mathbb{P}(y\in\mathcal{P}(x))\geq 1-\alpha,
    \label{eq:coverage}
\end{equation}
for a coverage level $\alpha \in (0, 1)$ specified by the user.
$\alpha$ is generally considered to represent a user pre-specified error rate.
For instance, if $\alpha$ is set to $0.1$, the resulting prediction set is expected to achieve a $90\%$ coverage rate. \textit{In other words, there is a $90\%$ probability that the true label will be included within the prediction set.}

However, the coverage in Eq.~(\ref{eq:coverage}) is guaranteed only when the testing domains are with the same distribution with the training domain, say data exchangeability~\cite{vovk2005algorithmic,barber2023conformal,zaffran2022adaptive,bhatnagar2023improved,gibbs2022conformal,farinhas2023non,zou2024coverage}.
When the domain shifts, the exchangeability is not satisfied, thus the coverage will significantly drop.
As observed by previous works~\cite{yilmaz2022test,bhatnagar2023improved}, even subtle shifts make coverage drop decline sharply.
This phenomenon is called \textbf{coverage gap}~\cite{barber2023conformal}, which is defined as follows:
\begin{equation}
    \kappa=(1-\alpha)-\mathbb{P}\left(y\in\mc{P}(x)\right),
\end{equation}
where $1-\alpha$ is the expected coverage and $\mathbb{P}\left(y\in\mc{P}(x)\right)$ is the obtained coverage.
To fill in the coverage gap, NexCP~\cite{barber2023conformal} generalizes CP by employing weighted quantiles and a randomization technique, enabling robust predictive inference even when data exchangeability assumptions are violated.
However, this method is designed for training phase and highly depends on a pre-defined domain shift value, which is not allowed in testing time.
Moreover, QTC~\cite{yilmaz2022test} recalibrate the quantile for coverage compensation. 
{However, QTC suffers from the unreliable domain gap measurement in continual domain shifts and ignores the model differences.}
More details about existing non-exchangeable CP can be found in Section \ref{sec:compareCP}.

Motivated by this, in this paper, we seek to design a CP method for CTTA to act as an uncertainty indicator during testing time, and solve the coverage gap issue.
Moreover, we would present to improve the adaptation in CTTA via the uncertainty measurement.

\section{Conformal Uncertainty Indicator for CTTA}

\subsection{CP with quantile compensation}
\label{sec:cui}

In this section, we propose a simple uncertainty indicator for the CTTA task named Conformal Uncertainty Indicator (CUI).
CUI is based on CP, and the major challenge of CUI is the coverage gap when the domain shifts.
In the following, we introduce how to build a CUI for CTTA task step by step.

\subsubsection{Step 1: Two ways to prepare calibration set}

First, following~\cite{vovk2005algorithmic}, CP needs to build a calibration set to approximate the source distribution for efficient computation.
The calibration set is with a small number of labeled samples drawn from the same distribution of the training data.
The calibration set is used to approximate the source domain distribution.
For different scene requirements, we can adopt two different calibration set construction methods, namely privacy-first and efficiency-first.

\noindent
\textit{(1) Privacy first}: The calibration set must not overlap with the training set. This scenario is common in traditional TTA tasks, where it is assumed that training data cannot appear during testing. In such cases, additional data with the same distribution as the training data can be collected to serve as the calibration set. Alternatively, a small portion of the existing training set can be split off to act as the calibration set before pretraining, with the pretraining process conducted only on the data excluding the calibration set. This paper adopts the latter approach, retraining the pre-trained model.

\noindent
\textit{(2) Efficiency first}: To make the most use of existing training data and pre-trained models, meaning the calibration set is sampled from the training data.
We select a part of labeled source data as the calibration set in our implementation.

{We will discuss the storage of calibration set construction in the end of the section.}
Specifically, we denote the calibration set as $\mc{C}=\{(x_1,y_1)\cdots,(x_{|\mc{C}|},y_{|\mc{C}|})\}$.
The calibration set should be built before the test phase.
Note that our method is only applied to CTTA tasks with this prepared calibration set, where the calibration data can be regarded as a fixed clue of training distribution.

\subsubsection{Step 2: Computing joint domain shifts}

Existing non-exchangeable CP methods fail to estimate the continual domain shifts in CTTA, such as NexCP~\cite{barber2023conformal} and QTC~\cite{yilmaz2022test}.
These methods either assume that the domain shift is known or ignore the issue of error accumulation.
In many existing domain difference measure methods, they directly compute distribution distance based on the current model.
For example, DSS~\cite{chakrabarty2023simple} uses the cosine distance between the prototypes of source domain and the current domain as the signal of domain shifts.
However, because the error accumulation, the current model could be not convincing enough.
That is, the prototypes may not represent the real data distributions.
To this end, we propose to further consider the \textit{model shift} when measure the domain shifts.

In our method, to estimate the domain shifts during continuous test time, we consider both model and data difference.
For model difference, we use both the source model $\boldsymbol{\theta}^\text{src}$ and the current model $\boldsymbol{\theta}^\text{crt}$.
For data difference, we use both the calibration set $\mc{C}$ and the current test data $\mc{B}$.
Specifically, we construct a joint probability distribution of calibration data and test data from both source and current models.
The joint probability distribution is computed by
\begin{equation}
    p(x) = \text{softmax}\left(\text{concat}(\pi_{\boldsymbol{\theta}^\text{src}}(x),\pi_{\boldsymbol{\theta}^\text{crt}}(x))\right).
\end{equation}
In this way, each sample can be represented by both the source and current models.
Then, for the joint distribution difference measurement, we use
\begin{equation}
\rho=\sum_{x^\text{calib}\in\mc{C}}\sum_{x^\text{test}\in\mc{B}} D_{\text{JS}}(p(x^\text{test})||p(x^\text{calib})),
    \label{eq:rho}
\end{equation}
where $D_{\text{JS}} $ is the Jensen-Shannon (JS) divergence, which is known as symmetric and stable. 
In the context of CTTA, joint feature representation captures correlations between different features, providing a more holistic view of the data distribution and how different models process it.
By combining multiple features, the joint distribution can better reflect subtle differences between domains, enhancing the precision of JS divergence measures.
Moreover, comparing joint feature distributions allows for a more detailed assessment of how much the current model has gained compared to the source model.

\subsubsection{Step 3: Compensating quantile threshold}

When obtaining the domain shift score $\rho$, we can compensate the coverage of CP in CTTA.
Specifically, we use the threshold conformal predictor (THR, ~\cite{sadinle2019least}) to
construct the prediction sets by thresholding output.
In general, the prediction set for the test sample $x$, denoted as $\mc{P}(x;\tau)$, are defined as the set of indices where the non-conformity score are greater than or equal to a threshold value $\tau$. 
The threshold value $\tau$ is determined as the ${(1-\alpha)}(\frac{|\mc{C}|+1}{|\mc{C}|})$-quantile of the calibrated non-conformity scores:
\begin{equation}
% \begin{aligned}
    \tau^* = \mathrm{Quantile}(\mc{C}, (1-\alpha)) 
    = \inf\left\{\tau:\mbb{E}_{{x}\in\mathcal{C}}\mathbb{I}_{\{s(\pi({x}))<\tau\}}\geq {\frac{|\mc{C}|+1}{|\mc{C}|}}(1-\alpha) \right\},
% \end{aligned}
\end{equation}
where the non-conformity scores $s(\cdot)$ represent the threshold required for each calibration example to achieve coverage, and can be easily computed by one minus the softmax output of the true class for the calibrated data:
\begin{equation}
    s(\pi_{\boldsymbol{\theta}^\text{crt}}(x))= 1 - \hat{y}.
    \label{eq:conf_score}
\end{equation}
Finally, we compensate the threshold based on the computed domain shift estimation $\rho$ in Eq.~(\ref{eq:rho}):
\begin{equation}
    \hat{\tau} = \tau^* - \beta\cdot\rho,
    \label{eq:comp}
\end{equation}
where $\beta$ is a predefined factor.
The compensation can be seen to include some uncertain classes to the prediction set to meet the coverage requirement.

\subsubsection{Step 4: Computing the prediction set} 

With the compensated threshold, we can compute the corresponding prediction set for the test sample $x$ by thresholding
\begin{equation}
    \mc{P}(x;\hat{\tau}) = \{y | s(y|\pi(x))<\hat{\tau}, \forall y \in \mc{Y}^{\mathrm{test}}\},
\end{equation}
where $\mc{Y}^{\mathrm{test}}$ is the label space.
In CP, the size of the prediction set can be seen as the measurement of uncertainty.
Generally, a prediction set with a large size is regarded as uncertainty.
The CUI algorithm can be seen in Algorithm~\ref{alg:CUI}.

\renewcommand{\algorithmicrequire}{\textbf{Input:}}
\renewcommand{\algorithmicensure}{\textbf{Output:}}

\begin{algorithm}[t]
\caption{{Conformal Uncertainty Indicator in CTTA}}
\label{alg:CUI}
\begin{algorithmic}[1] %[1] enables line numbers
\REQUIRE Test data point $x$, Pre-trained model $\pi$, calibration set $\mc{C}$, test data stream $\mc{X}^\text{test}$
\STATE Point prediction via the pre-tained model: $\hat{y} =\arg\max_{y\in\mathcal{Y}}{\pi}({y|x})$
\STATE Measure domain difference $\rho$ using Eq.~(\ref{eq:rho})
\STATE Compute non-conformity scores for calibration set using Eq.~(\ref{eq:conf_score})
\STATE Obtain the threshold $\tau^* = \text{Quantile}(\mc{C},1-\alpha) $ 
\STATE Compensate threshold via $\hat{\tau} = \tau^* - \beta\cdot\rho$
\STATE Set prediction via threshold: $\mathcal{P}(x;\hat{\tau})=\{y|s(y|\pi(x))<\hat{\tau},\forall y\in\mathcal{Y}\}$
\ENSURE Point prediction $\hat{y}$, Set prediction $\mc{P}$
\end{algorithmic}
\end{algorithm}

\subsection{CUI-guided adaptation}

\label{sec:cpada}

The size of the prediction set for each test sample represents the uncertainty level of the prediction.
The set size is close to $1$ but larger than $0$ can be regarded to reliable.
However, traditional CP methods focus on detecting violations of the exchangeability assumption rather than adapting to such changes~\cite{fedorova2012plug,volkhonskiy2017inductive,vovk2020computationally}.
In the context of CTTA, we prefer to further improve the adaptations via the guidance from CP.

Motivated by this, we design a simple adaptation strategy for CTTA based on CUI, weighting the adaptation of each test sample according to its prediction uncertainty. %say the size of the prediction set.
A test sample with a more reliable prediction will be set to a larger weight for adaptation.
Taking the adaptation in Mean-Teacher-based methods~\cite{wang2022continual,brahma2023probabilistic} as an example, where a student model updates via learning logits from a teacher, and the teacher then updates via exponential moving averaging (EMA) from the updated student.
In this case, the CUI-guided adaptation on the student model can be represented by:
\begin{equation}
  L = -\mbb{E}_{x\in\mathcal{B}}\gamma(x)\cdot\pi_{\boldsymbol{\theta}^\text{tea}}(x) \log \pi_{\boldsymbol{\theta}^\text{stu}}(x),
\end{equation}
where ${\boldsymbol{\theta}^\text{tea}}$ and ${{\boldsymbol{\theta}^\text{stu}}}$ are the teacher and student models, respectively.
$\gamma\left[x,\mc{P}(\mc{B};\tau)\right]$ is a function to assign weight to each adaptation and is highly related to the prediction set size:
\begin{equation}
    \gamma(x) = \left\{
    \begin{aligned}
        \frac{\max_{x'\in\mc{B}}(|\mc{P}(x')|)-|\mc{P}(x)|+\delta}{\max_{x'\in\mc{B}}(|\mc{P}(x')|)-1+\delta}, \quad &\text{if}~~|\mc{P}(x)|>0 \\
        0, \quad &\text{if}~~|\mc{P}(x)|=0,
    \end{aligned}
    \right.
    \label{eq:gamma}
\end{equation}
where $\mc{P}(x)=\mc{P}(x;\tau)$ for simplicity and $\delta$ is a minimum value (like $1e-9$) to avoid zero denominator.
Eq.~(\ref{eq:gamma}) gives a simple relative weight for a mini-batch adaptation.
Note that if the prediction set size is 1, \ie, $|\mc{P}(x)|=1$, we have $\gamma= 1$, which is considered as the most reliable.
Moreover, if $|\mc{P}(x)|=0$, that means an empty prediction set, we set the most unreliable prediction across the mini-batch.

\subsection{Discussion}

\subsubsection{Comparison with existing non-exchangeable CP methods} 
\label{sec:compareCP}

We compare our CUI with two recent non-exchangeable CP methods, including NexCP~\cite{farinhas2023non} and QTC~\cite{yilmaz2022test}.
First, both NexCP and QTC are designed only for uncertainty indication instead of adaptation improvement.
NexCP is designed for training time, where it specifies a constant to represent the domain difference from the source domain to the target domain.
Specifically, NexCP directly compensates the coverage by
\begin{equation}
    \mathbb{P}(y\in\mathcal{P}(x))\geq1-\alpha-2\sum_{i=1}^n{w}_i\epsilon_i,
\end{equation}
where $\epsilon_i$ is a predefined constant measure of how much the distribution has shifted from the test sample to the $i$-th calibrated sample 
and ${w}_i$ is a corresponding weight.
NexCP will satisfy marginal coverage, and are exact when the magnitude of the distribution shift is known, which is infeasible in test time.
In contrast, CUI is designed for testing, and measuring the distribution shifts adaptatively.

QTC proposes to replace the user-specified $\alpha$ to a new coverage level $\beta_{\mathrm{QTC}}$ calculated as
\begin{equation}
% \begin{aligned}
    \beta_{\mathrm{QTC}}=\min\big[\mbb{E}_{{x}\in\mathcal{C}}\mathbb{I}_{\{s(\pi({x}))<\mathrm{Quantile}(\mathcal{B},\alpha)\}},
    1-\mbb{E}_{{x}\in\mathcal{B}}\mathbb{I}_{\{s(\pi({x}))<\mathrm{Quantile}(\mathcal{C},1-\alpha)\}}
    \big].
% \end{aligned}
\end{equation}
Based on the current model $\pi$, QTC finds a threshold on the scores of the model on the unlabeled samples and predicts the coverage level by utilizing how the distribution of the scores changes across test distribution with respect to this threshold.
However, QTC ignore the adaptation on continual domain shifts may suffer serious error accumulation, making the current model unreliable.
This leads to the CP results being unreliable too.
Instead, our CUI considers the error accumulation and evaluates domain shifts based on a joint distribution difference.
% More details are shown in Appendix.

\subsubsection{Data storage for calibration in testing time} 
\label{sec:store}

In our method, we explore the feasibility of using CP in testing scenarios with the aid of additional samples for calibration.
That means the testing system needs to store extra data, yielding more storage requirements.
This is a common practice in continual learning.
Many continual learning~\cite{rolnick2019experience,van2020brain,lyu2021multi,sun2022exploring} methods store and retrain previous training examples to avoid catastrophic forgetting of past tasks, named replay strategy.
In comparison with replay, the calibration set in CUI is not used for adaptation but calibration in testing time, and the calibration set will not be updated in our method.
Practical approaches in real-world settings involve storing samples to improve testing outcomes, 
such as \cite{tomani2021post} and \cite{rahimi2020post} leverage post-hoc calibration to achieve better performance under domain drift scenarios by using validation or calibration sets.
In the CTTA tasks, some existing methods use source data to improve the adaptation such as \cite{dobler2023robust}.

The proposed CUI is plug-and-play, particularly well-suited for scenarios where the continuous accumulation of errors over long-term testing periods is unacceptable, such as in autonomous driving and medical applications. In these contexts, proactively assessing model uncertainty is essential to ensure safety and reliability, and it is acceptable for users to maintain a small set of calibration data to enhance the model’s performance and dependability. 
Furthermore, for a fair comparison, calibration sets are consistently employed across all methods discussed in the experiments.

\section{Experiment}
\label{sec:ex}

\subsection{Experimental Setting}

\textbf{Dataset}.
In our experiments, we employ the CIFAR10-to-CIFAR10C, CIFAR100-to-CIFAR100C, and ImageNet-to-ImageNetC datasets as benchmarks to assess the effectiveness of CUI (CIFAR10C, CIFAR100C and ImageNetC for short). Each dataset comprises 15 distinct types of corruption, each applied at five different levels of severity (from 1 to 5). These corruptions are systematically applied to test images from the original CIFAR10 and CIFAR100 datasets, as well as validation images from the original ImageNet dataset.
% For simplicity in tables, we use C1 to C15 to represent the 15 types of corruption, \ie,
% C1: Gaussian, C2: Shot, C3: Impulse C4: Defocus, C5: Glass, C6: Motion, C7: Zoom, C8: Snow, C9: Frost, C10: Fog, C11: Brightness, C12: Contrast, C13: Elastic, C14: Pixelate, C15: Jpeg.

\noindent
\textbf{Pretrained Model}.
Following previous studies~\cite{wang2020tent,wang2022continual}, we adopt pretrained WideResNet-28~\cite{zagoruyko2016wide} model for CIFAR10C, pretrained ResNeXt-29~\cite{xie2017aggregated} for CIFAR100C, and standard pretrained ResNet-50~\cite{he2016deep} for ImagenetC. 
Similarly, we update all the trainable parameters in all experiments. 
The augmentation number is set to 32 for all methods that use the augmentation strategy.
For a fair comparison, we conduct all experiments in a same environment.

\noindent
\textbf{Evaluation Metric}: 
We use two kinds of metrics including testing performance, CP performance.
We use $\hat{\mc{D}}$ to represent the testing data with labels.
(1) For testing performance, we use the error rate (ERR) following existing CTTA methods~\cite{wang2022continual}.
(2) For CP performance, we leverage coverage and inefficiency for joint evaluation:
\begin{equation*}
    % \small
    \mathrm{COV}=\mathbb{E}_{(x, y)\in \hat{\mc{D}}} \mbb{I} \left({y}\in \mc{P}(x)\right),~
    \mathrm{INE}=\mathbb{E}_{x\in \hat{\mc{D}}}\left|\mc{P}(x)\right|.
\end{equation*}
The coverage should be near to the user expectation and the inefficiency should be small but larger than 0.
Specifically, COV closer to $1-\alpha$ indicates a more effective uncertainty estimation of the CP. For example, with $\alpha = 0.1$, the COV should be close to $90\%$. INE, on the other hand, indicates lower uncertainty when closer to 1, while values closer to 0 suggest that no valid prediction. INE greater than 1, with larger values indicating higher uncertainty.

\subsection{Major Results}

\noindent
\textbf{Baseline Methods}: 
CUI is a play-and-plug uncertainty indicator. 
{To evaluate the effect of CUI, we select several well-known and state-of-the-art methods as the baseline methods.}
TENT~\cite{wang2020tent} updates via Shannon entropy for unlabeled test data.
CoTTA~\cite{wang2022continual} builds the MT structure and uses randomly restoring parameters to the source model. 
SATA~\cite{chakrabarty2023sata} modifies the batch-norm affine parameters using source anchoring-based self-distillation to ensure the model incorporates knowledge of newly encountered domains while avoiding catastrophic forgetting.
RMT~\cite{dobler2023robust} combines symmetric cross-entropy with contrastive learning in CTTA.
C-CoTTA~\cite{shi2024controllable} proposes to adjust the directions of domain shift therefore to keep the discriminative ability.
% For each selected method, we have two parts of results that are shown as two adjacent rows, such as CoTTA and CoTTA + CPAda, where the first 
RDumb~\cite{press2024rdumb} proposes to evaluate the asymptotic performance in CTTA, and reset the model to its pre-trained state periodically to avoid performance collapse.

\noindent
\textbf{Implementation Details}: 
We set the total calibration set sizes to 50, 100, and 500 for CIFAR10C, CIFAR100C, and ImageNetC, respectively.
All compared methods adopt the same pre-trained model under the same calibration set construction strategy, which can be privacy first or efficiency first.
For each selected method, we use the proposed CUI for uncertainty measurement, and based on this, we compare two results: one without adaptation and one using CUI guidance for domain adaptation.
These two results are represented as adjacent rows in the table, such as ``CoTTA+\textcolor{red!60}{CUI}'' and ``CoTTA+\textcolor{red!60}{CUI}+\textcolor{blue!60}{CPAda}''.
We use three expected coverage factors $\alpha=0.1, 0.2, 0.3$, which represents that the user would like $90\%, 80\%, 70\%$ coverage for the prediction.

\begin{table*}[t]
\caption{Results of combining CUI with exiting CTTA methods on the three datasets. All results are evaluated with the largest corruption severity level 5 in an online fashion.
For each SOTA method, the first line means the vanilla implementation only with CUI for uncertainty estimation, and the second line means the method uses uncertainty to guide the adaptation.
\textit{Because CUI does not change the ERR, we omit the results of these methods w/o CUI and w/o CPAda for saving spac}e.
}
\label{tab:cifar10c}
% \vskip -0.25in
\vspace{-10px}
\begin{center}
\resizebox{\linewidth}{!}{
\begin{tabular}{l|l||ccc|ccc|ccc||ccc|ccc|ccc}
\toprule
&\textbf{Method} & \multicolumn{9}{c||}{\textbf{Privacy First} (Calibration data $\cap$ Training data $=\emptyset$)} & \multicolumn{9}{c}{\textbf{Efficiency First} (Calibration data $\subset$ Training data)} \\
&{\quad 1. \textbf{\color{red!60}CUI}: Sec.~\ref{sec:cui}}&\multicolumn{3}{c|}{$\alpha=0.3$}&\multicolumn{3}{c|}{$\alpha=0.2$}&\multicolumn{3}{c||}{$\alpha=0.1$}&\multicolumn{3}{c|}{$\alpha=0.3$}&\multicolumn{3}{c|}{$\alpha=0.2$}&\multicolumn{3}{c}{$\alpha=0.1$} \\
&{\quad 2. \textbf{\color{blue!60}CPAda}: Sec.~\ref{sec:cpada}} & \textbf{ERR} & \textbf{COV} & \textbf{INE} & \textbf{ERR} & \textbf{COV} & \textbf{INE} & \textbf{ERR} & \textbf{COV} & \textbf{INE} & \textbf{ERR} & \textbf{COV} & \textbf{INE} & \textbf{ERR} & \textbf{COV} & \textbf{INE} & \textbf{ERR} & \textbf{COV} & \textbf{INE} \\
\midrule
&Tent + \textbf{\color{red!60}CUI}    &21.65 & 69.12 & 0.89 &21.65 & 78.45 & 1.96 &21.65 & 87.93 & 2.33&20.45&68.55&0.81&20.45&77.88&1.02&20.45&87.67&1.57\\
&Tent + \textbf{\color{red!60}CUI} + \textbf{\color{blue!60}CPAda} &19.70 & 69.04 & 0.81 &19.25 & 76.73 & 1.02 &19.25 & 87.17&1.56&18.06&67.91&0.77&18.32&78.19&1.01&18.22&87.57&1.29\\
\rowcolor{gray!20} \cellcolor{white} &
CoTTA + \textbf{\color{red!60}CUI}   &16.34 & 68.77& 0.81 &16.34 & 78.45&	1.89 &16.34 & 87.85&1.66&16.22&67.86&1.15&16.22&75.36&1.09&16.22&89.35&1.90 \\
\rowcolor{gray!20} \cellcolor{white} &
CoTTA + \textbf{\color{red!60}CUI} + \textbf{\color{blue!60}CPAda} &15.73&68.77&0.81&15.75&77.93&1.03 &15.71&87.02&1.46&15.52&66.62&0.81&15.73&77.25&	1.00&15.65&88.53&1.61 \\
&SATA + \textbf{\color{red!60}CUI}    &16.31&68.25&0.75
 &16.31&77.78&0.95 &16.31&86.07&1.24  &16.13&68.28&0.84&16.13	&77.14&0.85
&16.13&85.61&1.09\\
&SATA + \textbf{\color{red!60}CUI} + \textbf{\color{blue!60}CPAda} &15.79&68.83&0.75
&15.76&76.68&0.89&15.72&86.97&1.30&15.59&67.94&0.73&15.56&78.49&0.92&15.60&88.68&1.24
\\
\rowcolor{gray!20} \cellcolor{white} &
RDumb + \textbf{\color{red!60}CUI}    &18.31&68.62&0.76&18.31&78.82&0.94 &18.31&85.60&1.15&17.63&68.37&0.76&17.63&77.87&0.91&17.63&86.23&1.17\\
\rowcolor{gray!20} \cellcolor{white} &
RDumb + \textbf{\color{red!60}CUI} + \textbf{\color{blue!60}CPAda} &16.73 & 73.55 & 0.83  &16.73 & 79.30 & 0.94 &16.81&86.41&1.18&16.23&68.30&0.74&16.31&76.63&0.87&16.33&84.38&1.09\\
&C-CoTTA +\textbf{\color{red!60}CUI}     &14.99&68.39&0.73&14.99&	78.42&	1.23&14.99	&86.92&	1.75&14.74	&66.16	&0.70 &14.74	&77.46	&0.87
&14.74	&87.52	&1.44 \\
&C-CoTTA +\textbf{\color{red!60}CUI} + \textbf{\color{blue!60}CPAda}  &14.75&	66.97&	0.72&14.72&77.10&1.14&14.76&86.42&1.55&14.32&68.82&	0.74 &14.38	&75.53	&0.85&14.33	&88.47&	1.64\\
\rowcolor{gray!20} \cellcolor{white} &
RMT + \textbf{\color{red!60}CUI}    &14.66&68.86&0.75&14.66&76.81&1.14&14.66&87.37&1.45&14.54&68.29&0.85&14.54&78.37	& 1.10 &14.54 & 89.06 & 1.50\\
\rowcolor{gray!20}\multirow{-12}{*}{\rotatebox[origin=c]{90}{\textbf{CIFAR10-CIFAR10C}}} \cellcolor{white} &
RMT + \textbf{\color{red!60}CUI} + \textbf{\color{blue!60}CPAda} &14.33	& 66.53 &	0.72  &14.36 &	78.04 &	1.22 &14.44 & 86.29 & 1.26
&14.28 & 69.17 & 0.83 &14.31 & 77.28 & 0.91 &14.25 & 86.58 & 1.70
\\
\midrule
&Tent + \textbf{\color{red!60}CUI}    &62.24&69.23&2.66& 62.24&78.50&4.44&62.24&87.24&11.19&60.93&69.04&17.32&60.93&77.15&27.97&60.93&84.63&35.52\\
&Tent + \textbf{\color{red!60}CUI} + \textbf{\color{blue!60}CPAda} &46.50&68.88&1.53&46.56&76.85&3.68&45.95&87.42&4.13&49.87&68.22&20.66&52.90&78.93&24.34&51.56&84.48&28.61\\
\rowcolor{gray!20} \cellcolor{white} &
CoTTA + \textbf{\color{red!60}CUI}   &36.41&68.08&1.86&36.41&77.01&2.82&36.41&87.31&4.96&32.50&66.59&2.42&32.50&78.39&5.11&32.50&88.68&11.58 \\
\rowcolor{gray!20} \cellcolor{white} &
CoTTA + \textbf{\color{red!60}CUI} + \textbf{\color{blue!60}CPAda} &32.11&67.81&1.69&32.16&79.33&3.34&32.31&89.64&9.69&30.93	&64.65&1.85&30.99&75.08&3.16&31.59&84.61&6.45 \\
&SATA + \textbf{\color{red!60}CUI}    &33.46&69.32&1.81&33.46&76.84&2.79&33.46&87.39&7.06&30.30&68.69&1.55&30.30&77.80&2.64&30.30&87.82&6.02
\\
&SATA + \textbf{\color{red!60}CUI} + \textbf{\color{blue!60}CPAda} &32.38&68.36&1.65&32.39&77.85&2.92&32.46&89.51&8.64&29.14&68.81&1.44&28.94&76.29&2.08&28.78&84.92&3.69\\
\rowcolor{gray!20} \cellcolor{white} &
RDumb + \textbf{\color{red!60}CUI}    &45.93&68.01&2.29&45.93&76.86&3.38&45.93&88.48&7.23&45.10&68.56&2.06&45.10&78.02&2.21&45.10&87.68&2.23
\\
\rowcolor{gray!20} \cellcolor{white} &
RDumb + \textbf{\color{red!60}CUI} + \textbf{\color{blue!60}CPAda} &42.12&68.62&1.76&42.23&79.30&2.94&42.26&86.21&7.89&43.42&69.49&2.72&43.22&76.10&2.86&43.36&85.28&3.40
\\
&C-CoTTA +\textbf{\color{red!60}CUI}     &32.79&68.58&1.83&32.79&78.12&3.21&32.79&88.37&7.62&29.90&69.75&1.71 &29.90&76.54&2.51&29.90&84.51&4.70\\
&C-CoTTA +\textbf{\color{red!60}CUI} + \textbf{\color{blue!60}CPAda}  &31.52&68.08&1.66&31.44&77.96&2.97&31.47&88.19&7.20&29.31&68.79&2.46 &29.28&78.64&2.60&29.17&86.08&5.32\\
\rowcolor{gray!20} \multirow{-12}{*}{\rotatebox[origin=c]{90}{\textbf{CIFAR100-CIFAR100C}}}\cellcolor{white} &
RMT + \textbf{\color{red!60}CUI}    &32.53& 68.37&1.45 &32.53 &77.06 &	2.75&32.53 & 88.48 & 7.46 &29.00 & 69.41 &	1.69 &29.00 &	76.71 &	2.62 &29.00&	87.97 &	5.80
\\
\rowcolor{gray!20} \cellcolor{white} &
RMT + \textbf{\color{red!60}CUI} + \textbf{\color{blue!60}CPAda} &31.43	&67.47 & 1.39 &31.32 &	76.71 &	2.62 &31.45 &	86.97 &	6.40 &28.35 &	67.67 &	1.40 &28.33 & 77.06 & 2.75 &28.28 &	87.71 &	4.49\\
\midrule
&Tent + \textbf{\color{red!60}CUI}    &63.69&68.12&43.09 &63.69&78.07&114.50&64.69&87.42&265.59&62.60&69.09&47.80&62.60&79.40&82.62&62.60&88.48&163.09
\\
&Tent + \textbf{\color{red!60}CUI} + \textbf{\color{blue!60}CPAda} &62.50&69.26&47.89&62.53&76.89&112.25&62.60&88.71&272.71&61.50&69.26&47.89&61.53&76.19&43.25&61.60&88.71&164.50
\\
\rowcolor{gray!20} \cellcolor{white} &
CoTTA + \textbf{\color{red!60}CUI}   &69.03&68.88&84.43&69.03&79.01&110.13&69.03&88.28&188.43&62.70&68.43&69.74&62.70&78.07&90.86&62.70&86.70&171.33 \\
\rowcolor{gray!20} \cellcolor{white} &
CoTTA + \textbf{\color{red!60}CUI} + \textbf{\color{blue!60}CPAda} &67.56&67.74&80.13&67.42&78.42&114.43&67.32&89.04&179.64&61.22&69.01&69.32&61.30&77.42&86.23&61.24&87.40&172.24 \\
&SATA + \textbf{\color{red!60}CUI}    &61.81&69.83&81.31&61.81&76.97&118.13
&61.81&87.95&212.59&60.10&69.38&75.93&60.10&77.42&120.44&60.10&88.12&218.29\\
&SATA + \textbf{\color{red!60}CUI} + \textbf{\color{blue!60}CPAda} &60.62&69.10&54.99&60.92&79.09&113.38&60.87&89.46&224.14&58.52&68.24&64.18&58.54&78.71&121.57&58.65&87.32&192.66\\
\rowcolor{gray!20} \cellcolor{white} &
RDumb + \textbf{\color{red!60}CUI}    &64.46&66.68&21.67&64.46&79.49&57.39&64.46&88.55&156.87&62.45 &	67.54 &	23.38&62.45 &	79.22 &	67.03&62.45 &88.83 &	147.44 \\
\rowcolor{gray!20} \cellcolor{white} &
RDumb + \textbf{\color{red!60}CUI} + \textbf{\color{blue!60}CPAda} &62.25 &67.29 &22.74&62.29 &78.28 &56.45 &62.18&87.34&152.11&60.26 &	67.57 &	24.52
&60.32 &	78.39 &	62.16&60.54 &	89.01	&156.74
\\
&C-CoTTA +\textbf{\color{red!60}CUI} &60.42&68.11&36.13&60.42&75.19&32.61&60.42&87.70&91.22&59.40&67.45&17.09&59.40&78.14&39.26&59.40&88.09&100.20
\\
&C-CoTTA +\textbf{\color{red!60}CUI} + \textbf{\color{blue!60}CPAda}  &59.48&68.03&20.87&59.52&77.24&42.90&59.53&88.74&96.05 &58.36&68.31&18.73&58.33&79.05&40.46&58.39&87.67&98.40
\\
\rowcolor{gray!20} \cellcolor{white} &
RMT + \textbf{\color{red!60}CUI}    &61.64&69.79&19.44&61.64 & 78.05&37.59&61.64&86.15&82.13&59.80&69.53&18.73&59.80&78.04&38.18&59.80&86.83&82.37
\\
\rowcolor{gray!20} \multirow{-12}{*}{\rotatebox[origin=c]{90}{\textbf{ImageNet-ImageNetC}}} \cellcolor{white} &
RMT + \textbf{\color{red!60}CUI} + \textbf{\color{blue!60}CPAda} &59.62 &69.71 &	18.98
&59.65 &	78.57 &	39.07 &59.66 &	86.04 &	76.99
 &59.28 &69.57 &	19.30 &59.25 & 76.91 &	34.27 &59.30 &	87.35 &87.60
\\
\bottomrule
\end{tabular}
}
\end{center}
% \vskip -0.2in
% \vspace{-15px}
\end{table*}

\noindent
\textbf{Observations and Analysis}: 
The results are shown in Tables~\ref{tab:cifar10c}, and the analysis reveals several key observations.
First, with the inclusion of CUI, it is possible to estimate uncertainty (INE) that closely aligns with the predefined $\alpha$ values.
In most cases, when CPAda is not employed, the INE values reveal significant inherent uncertainties within the baseline method itself.
These uncertainties are strongly associated with the dataset that more complex datasets typically exhibit higher INE values. 
Moreover, the INE varies depending on the $\alpha$ value. Specifically, smaller $\alpha$ values correspond to larger INE, as smaller $\alpha$ thresholds demand higher fault tolerance. This relationship highlights the trade-off between the level of certainty required and the algorithm’s ability to meet that requirement.
Second, the integration of CUI-guided CPAda improves existing methods, reducing error rates (ERR) and lowering INE, indicating more accurate and confident predictions. 
Finally, the comparison between Privacy-First and Efficiency-First strategies shows minimal performance differences, suggesting that users can select the calibration dataset construction method based on their specific application needs without compromising results.

\subsection{More analysis on the proposed method}

\subsubsection{Comparisons with non-exchangeable CP methods}

In Table~\ref{tab:cp_comapre}, we compare our CUI with other CP methods including THR~\cite{sadinle2019least}, NexCP~\cite{barber2023conformal} and QTC~\cite{yilmaz2022test}.
THR is an exchangeable CP method and never considers domain shifts in CTTA, thus it obtains an obvious coverage gap.
NexCP and QTC are two non-exchangeable methods, with detailed comparisons available in Sec.~\ref{sec:compareCP}. 
First, for NexCP, we use the same fixed value for domain shift estimation as in the original paper, and NexCP is only slightly better than THR and struggles to estimate domain differences in advance during testing.
Then, although QTC estimates domain differences in real time, it neglects the unreliability of the current model due to error accumulation over long testing periods. This method yields better results than both THR and NexCP. 
\textit{However, these methods all suffer from coverage gap issues, and the uncertainty estimation is unreliable in CTTA, even if their INE is close to 1.}
Instead, CUI obtains near-expected coverage when estimating testing uncertainty.
Next, we compare our domain adaptation method (CPAda) using different CP techniques that similar to the proposed method, and the results show that CUI provides better guidance for adaptation and obtains less error rates.

\begin{table}[h]
    \centering
    \setlength{\tabcolsep}{3pt}
    \caption{{Comparisons with non-exchangeable CP methods}.}
    \vspace{-5px}
    \resizebox{.6\linewidth}{!}{
    \begin{tabular}{cl||ccc|ccc||ccc|ccc}
    \toprule
        &&\multicolumn{6}{c||}{\textbf{Privacy First}} & \multicolumn{6}{c}{\textbf{Efficiency First}}\\
        &\textbf{CP}&\multicolumn{3}{c|}{w/o CPAda} & \multicolumn{3}{c||}{w/ CPAda} &\multicolumn{3}{c|}{w/o CPAda} & \multicolumn{3}{c}{w/ CPAda}\\
        $\alpha$ & \textbf{Method} & ERR & COV & INE & ERR & COV & INE& ERR & COV & INE& ERR & COV & INE \\
        \midrule
        & N/A & 35.15& 23.39 & 0.28 & - & - & - & 32.77 & 34.27 & 0.44 & -  & -  & - \\
        \midrule
        \multirow{4}{*}{0.3}&THR & 35.15 &	23.39 &	0.28 & 35.18 &	21.31 &	0.24 & 32.72 & 34.17 & 0.44 &31.89 &	39.81 &	0.50 \\
        &NexCP & 35.15 &	23.46 &	0.28 & 35.21 &	21.75 &	0.25  &32.72 &34.68 &0.45 & 31.70 &	40.31 &	0.51 \\
        &QTC & 35.15 &	40.70 &	0.59 & 33.79 &	42.25 &	0.59  &32.72 & 52.15 &0.87 & 31.00 &	53.13 &	0.75\\
        &CUI &35.15 &	69.64 &	2.70  & \textbf{32.76} &	68.02 &	2.18 &32.72 & 68.95 & 2.01 &\textbf{29.48} & 68.07 & 2.17 \\
        \midrule
        \multirow{4}{*}{0.2}&THR &35.15	& 29.05 &	0.37 &34.80 &	27.93 &	0.34 &32.72 & 42.17 & 0.60 &31.43 &	48.32 &	0.67 \\
        &NexCP & 35.15 &	29.42 &	0.37  &34.78 &	28.39 &	0.35  &32.72	& 41.87 & 0.59 &31.32 &	48.36	& 0.66 \\
        &QTC & 35.15 &	46.63 &	0.75  &33.54 &	47.84 &	0.73 & 32.72 &	59.96 &1.22 & 30.53 & 61.53 &	0.99\\
        &CUI &35.15 &	77.58	& 4.60 &\textbf{32.59} &	77.46 &	3.64 &32.72 & 76.73 &3.42 & \textbf{29.17} & 79.15 & 2.27\\
        \midrule
        \multirow{4}{*}{0.1}&THR &35.15 &	37.25 &	0.52  & 34.20 & 37.12 &	0.49 & 32.72 &	53.69 &	0.95 & 30.64 &	59.89 &	0.97\\
        &NexCP & 35.15 &	37.71 &	0.53 & 34.17 &	37.70 &	0.51 &32.72	& 53.17 &0.92 &30.62 &	59.83 &	0.97 \\
        &QTC & 35.15 &	55.56 &	1.10 & 33.25 &	54.29 &	0.93 &32.72 & 69.14 &1.92 & 29.58 &	72.31 &	1.50\\
        &CUI & 35.15 &	86.41 &	9.30 & \textbf{32.74} &	89.02 &	11.48  &32.72 &	86.38 &	7.78 & \textbf{29.17} &	88.35 &	5.47\\
     \bottomrule
    \end{tabular}}
    \label{tab:cp_comapre}
    % \vspace{-15px}
\end{table}

\begin{figure}[t]
    \centering
     \includegraphics[width=.7\linewidth]{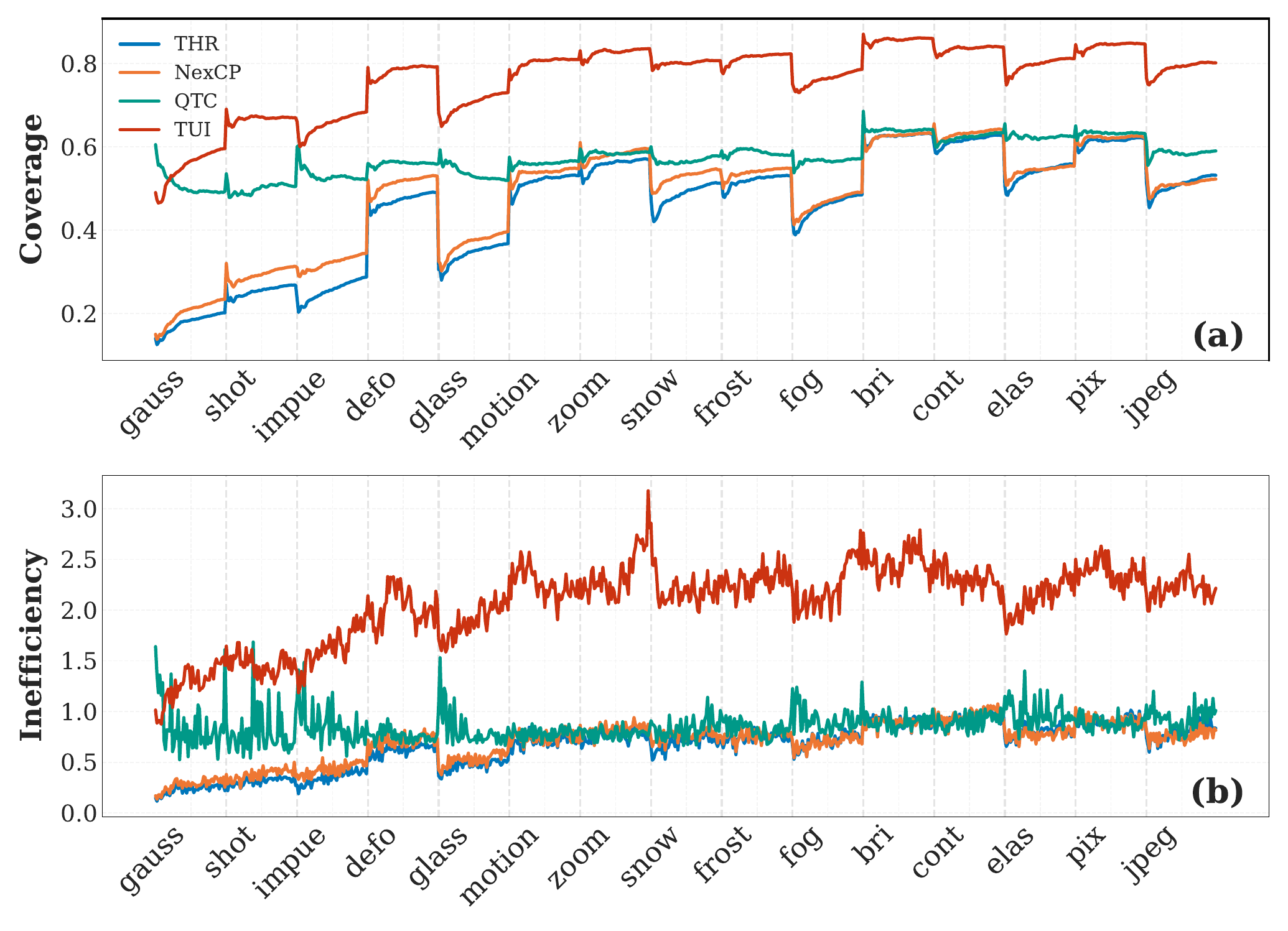}
    % \vspace{-20px}
    \caption{Visualization of coverage and inefficiency changes.}
    \label{fig:vis}
    % \vspace{-10px}
\end{figure}

% \subsubsection{Coverage and Inefficiency changes in CTTA}

In Fig.~\ref{fig:vis}, we show the coverage and inefficiency changes of different CP methods.
As shown in Fig.\ref{fig:vis}(a), coverage varies significantly across methods, reflecting domain disparities. Existing methods, such as THR and NexCP, show notable coverage gaps, while QTC performs well initially but struggles with error accumulation. In contrast, CUI achieves comparable initial coverage to QTC and surpasses it in later domains. Fig.\ref{fig:vis}(b) illustrates inefficiency trends, revealing that existing methods, despite low coverage, fail to account for error accumulation during domain shifts, leading to overconfidence. CUI, however, captures this accumulation, with inefficiency increasing as domains change, reflecting growing uncertainty. When CUI guides domain adaptation, inefficiency decreases, demonstrating effective uncertainty control.

\begin{table}[h]
    \centering
    \caption{{Storage analysis and comparison with replay strategy}.}
    % \vspace{-10px}
    \resizebox{.6\linewidth}{!}{
    \begin{tabular}{l|c|ccc||ccc}
    \toprule
        &&\multicolumn{3}{c||}{\textbf{Privacy First}} & \multicolumn{3}{c}{\textbf{Efficiency First}}\\
        \textbf{Method} & \textbf{Total Storage} & \textbf{ERR} & \textbf{COV} & \textbf{INE} & \textbf{ERR} & \textbf{COV} & \textbf{INE}\\
        \midrule
        % CE ~\cite{wang2022continual}  & - &  & &  \\
        Baseline & & 35.15 & 23.39 & 0.28 & 32.77 & 34.27 & 0.44  \\
        \midrule
        Soure Replay & & 35.03 & 21.88 & 0.25 & 32.64 & 7.47 & 0.08 \\
        CUI+CPAda & \multirow{-2}{*}{100}  & 32.59 & 78.11 &3.29 &29.17 & 79.15 & 2.27\\
        \midrule
        Soure Replay & &35.02 & 13.34 &	0.14 &   32.52 & 8.20 & 0.09 \\
        CUI+CPAda & \multirow{-2}{*}{200} & 31.97 &79.02&7.61 & 29.38 & 77.05 & 2.04 \\
        \midrule
        Soure Replay & & 34.22 &13.74 &	0.15 & 32.09 &	8.97 &	0.09 \\
        CUI+CPAda & \multirow{-2}{*}{300} & 31.33 &78.59 &5.12 & 29.77 & 77.48 & 2.18 \\
     \bottomrule
    \end{tabular}}
    \label{tab:replay}
    \vspace{-10px}
\end{table}

\subsubsection{Storage analysis and comparison with replay strategy}

As discussed in Sec.~\ref{sec:store}, CP-based methods need to maintain an extra calibration set for uncertainty estimation.
Although effectively measuring uncertainty is crucial in testing systems, using CP requires a certain amount of memory storage.
We analyze the impact of this storage on performance in Table~\ref{tab:replay} and find that a larger storage capacity leads to better CP performance, as more calibration data provides a more accurate representation of the original data distribution. 
Additionally, we compare CUI with a classic storage method in continual learning, the source replay strategy, where we use the same samples for replay when conducting adaptation.
We find that CUI achieves better accuracy while maintaining the same amount of stored data, which shows the significance of reducing error accumulation in CTTA.

\subsubsection{{Impacts of user-specified coverage level $\alpha$}}

In CP, we have a user specified coverage level $\alpha\in (0, 1)$ (Eq.~\eqref{eq:coverage}), which is generally considered to represent a user pre-specified error rate.
In Fig.~\ref{fig:hyperparameter}(a), we show that the infuence of different $\alpha$ from $0.1$ to $0.9$.
The results show that large $\alpha$ means that the user accept less coverage rate, reflecting large error rate.

\subsubsection{{Analysis of compensation factor $\beta$}}
\begin{figure}[t]
    \centering
    \includegraphics[width=.8\linewidth]{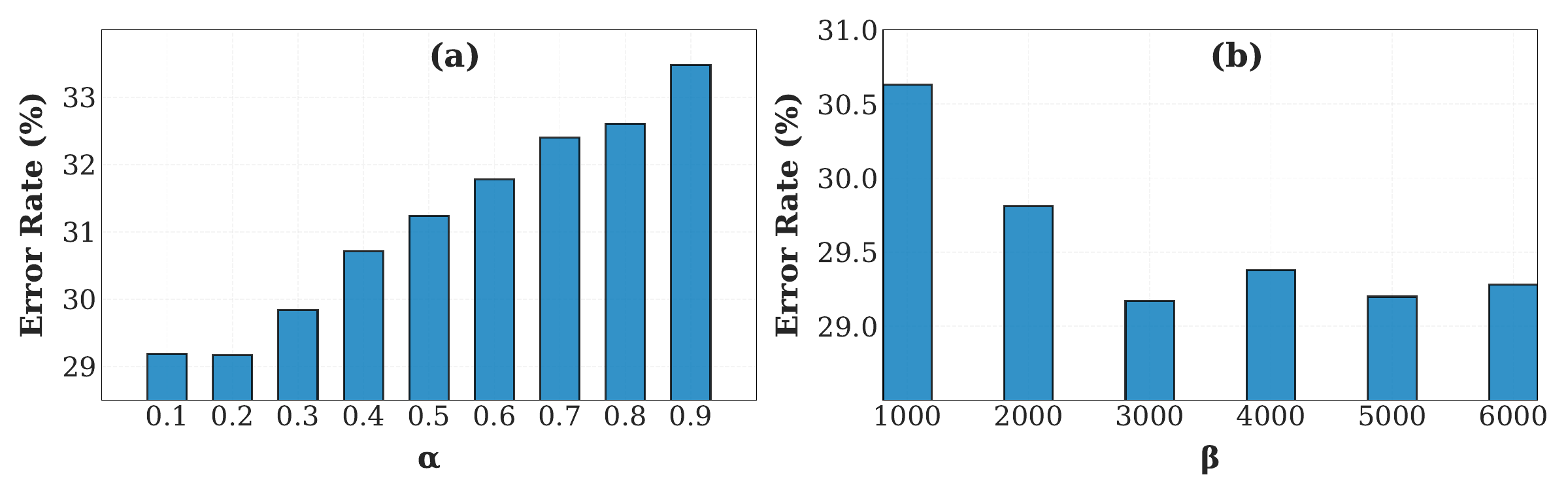}
    % \vspace{-15px}
    \caption{{Hyperparameter analysis on CIFAR100-to-CIFAR100C}.}
    \label{fig:hyperparameter}
    % \vspace{-10px}
\end{figure}

We also analysis the influence of different compensation factor $\beta$ in Eq.~\eqref{eq:comp}, which represents the compensation level.
The results are shown in Fig.~\ref{fig:hyperparameter}(b), we find that small $\beta$ decrease the compensation performance and large $\beta$ may result in overcompensation.

\begin{figure}[h]
    \centering
    \includegraphics[width=.8\linewidth]{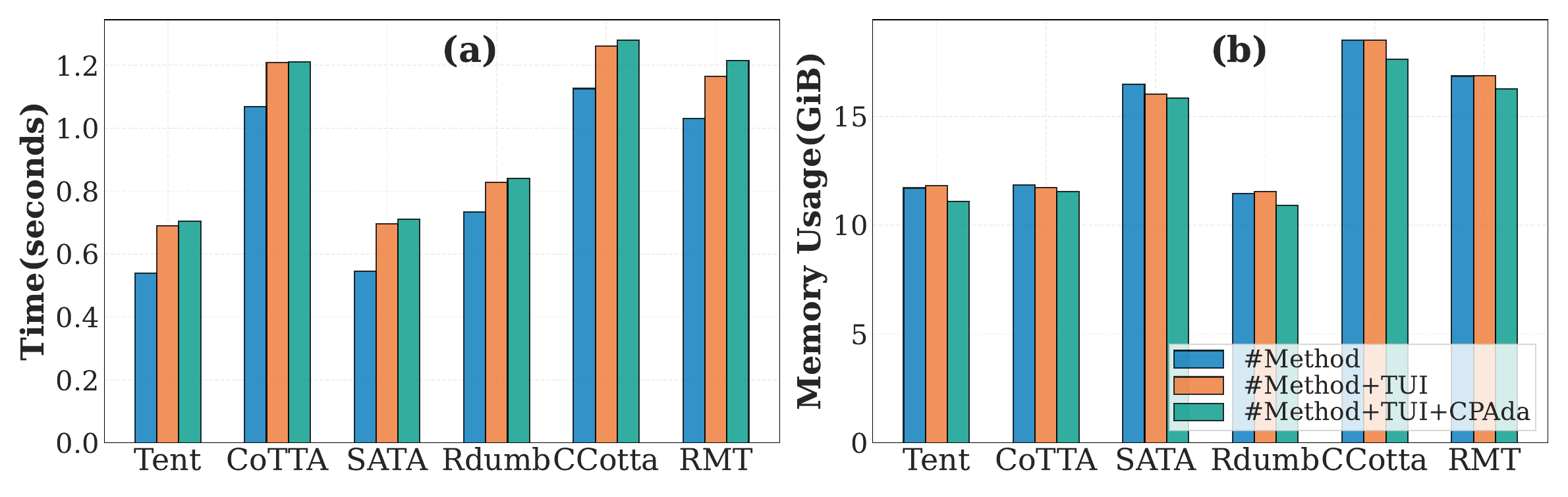}
    % \vspace{-15px}
    \caption{Time and memory cost on CIFAR100-to-CIFAR100C.}
    \label{fig:time}
    % \vspace{-10px}
\end{figure}

\subsubsection{{Time and memory cost}}

Since CUI is a plug-and-play module, we analyze its impact on time and memory costs compared to the original methods, as shown in Fig.~\ref{fig:time}. It is evident that our CUI and CPAda strategies slightly increase implementation time due to the forward propagation of calibration data. However, CPAda reduces memory costs by performing backpropagation only on selected samples.

\section{Conclusion}

CTTA is prone to error accumulation, where incorrect pseudo-labels can harm subsequent model adaptation. To address this, we propose CUI, a simple uncertainty indicator for CTTA based on CP, which generates a set of possible labels for each instance, ensuring the true label is included with a specified coverage probability. To reduce coverage gaps during domain shifting, we dynamically measure domain differences in continuously changing environments and use relabeled pseudo-labels to enhance adaptation. Experimental results show that our method effectively estimates uncertainty and improves adaptation performance across various CTTA methods.
However, CUI has two limitations: it requires a calibration set for conformal calculation, which may not always be available, and it only provides data-level uncertainty, limiting its application to tasks like pixel-level uncertainty in semantic segmentation. Future work will address these limitations and explore practical applications of CUI.

\bibliographystyle{plain}
\bibliography{references}  %%% Uncomment this line and comment out the ``thebibliography'' section below to use the external .bib file (using bibtex) .

\begin{thebibliography}{10}

\bibitem{barber2023conformal}
Rina~Foygel Barber, Emmanuel~J Candes, Aaditya Ramdas, and Ryan~J Tibshirani.
\newblock Conformal prediction beyond exchangeability.
\newblock {\em The Annals of Statistics}, 51(2):816--845, 2023.

\bibitem{bhatnagar2023improved}
Aadyot Bhatnagar, Huan Wang, Caiming Xiong, and Yu~Bai.
\newblock Improved online conformal prediction via strongly adaptive online
  learning.
\newblock In {\em International Conference on Machine Learning}, pages
  2337--2363, 2023.

\bibitem{brahma2023probabilistic}
Dhanajit Brahma and Piyush Rai.
\newblock A probabilistic framework for lifelong test-time adaptation.
\newblock In {\em Proceedings of the Computer Vision and Pattern Recognition},
  2023.

\bibitem{caruana2015intelligible}
Rich Caruana, Yin Lou, Johannes Gehrke, Paul Koch, Marc Sturm, and Noemie
  Elhadad.
\newblock Intelligible models for healthcare: Predicting pneumonia risk and
  hospital 30-day readmission.
\newblock In {\em Proceedings of the ACM SIGKDD international conference on
  knowledge discovery and data mining}, pages 1721--1730, 2015.

\bibitem{chakrabarty2023sata}
Goirik Chakrabarty, Manogna Sreenivas, and Soma Biswas.
\newblock Sata: Source anchoring and target alignment network for continual
  test time adaptation.
\newblock {\em arXiv preprint arXiv:2304.10113}, 2023.

\bibitem{chakrabarty2023simple}
Goirik Chakrabarty, Manogna Sreenivas, and Soma Biswas.
\newblock A simple signal for domain shift.
\newblock In {\em Proceedings of the IEEE/CVF International Conference on
  Computer Vision}, pages 3577--3584, 2023.

\bibitem{chen2024each}
Ziyang Chen, Yongsheng Pan, Yiwen Ye, Mengkang Lu, and Yong Xia.
\newblock Each test image deserves a specific prompt: Continual test-time
  adaptation for 2d medical image segmentation.
\newblock In {\em Proceedings of the IEEE/CVF Conference on Computer Vision and
  Pattern Recognition}, pages 11184--11193, 2024.

\bibitem{deng2009imagenet}
Jia Deng, Wei Dong, Richard Socher, Li-Jia Li, Kai Li, and Li~Fei-Fei.
\newblock Imagenet: A large-scale hierarchical image database.
\newblock In {\em IEEE/CVF Conference on Computer Vision and Pattern
  Recognition}, 2009.

\bibitem{dobler2023robust}
Mario D{\"o}bler, Robert~A Marsden, and Bin Yang.
\newblock Robust mean teacher for continual and gradual test-time adaptation.
\newblock In {\em Proceedings of the Computer Vision and Pattern Recognition},
  2023.

\bibitem{farinhas2023non}
Ant{\'o}nio Farinhas, Chrysoula Zerva, Dennis Ulmer, and Andr{\'e}~FT Martins.
\newblock Non-exchangeable conformal risk control.
\newblock {\em arXiv preprint arXiv:2310.01262}, 2023.

\bibitem{fedorova2012plug}
Valentina Fedorova, Alex Gammerman, Ilia Nouretdinov, and Vladimir Vovk.
\newblock Plug-in martingales for testing exchangeability on-line.
\newblock In {\em Proceedings of the International Conference on International
  Conference on Machine Learning}, pages 923--930, 2012.

\bibitem{gal2016dropout}
Yarin Gal and Zoubin Ghahramani.
\newblock Dropout as a bayesian approximation: Representing model uncertainty
  in deep learning.
\newblock In {\em International conference on Machine Ltoearning}, pages
  1050--1059. PMLR, 2016.

\bibitem{gibbs2022conformal}
Isaac Gibbs and Emmanuel Cand{\`e}s.
\newblock Conformal inference for online prediction with arbitrary distribution
  shifts.
\newblock {\em arXiv preprint arXiv:2208.08401}, 2022.

\bibitem{he2016deep}
Kaiming He, Xiangyu Zhang, Shaoqing Ren, and Jian Sun.
\newblock Deep residual learning for image recognition.
\newblock In {\em Proceedings of the Computer Vision and Pattern Recognition},
  2016.

\bibitem{jain2011online}
Vidit Jain and Erik Learned-Miller.
\newblock Online domain adaptation of a pre-trained cascade of classifiers.
\newblock In {\em Proceedings of the Computer Vision and Pattern Recognition},
  2011.

\bibitem{lekeufack2023conformal}
Jordan Lekeufack, Anastasios~A Angelopoulos, Andrea Bajcsy, Michael~I Jordan,
  and Jitendra Malik.
\newblock Conformal decision theory: Safe autonomous decisions from imperfect
  predictions.
\newblock {\em arXiv preprint arXiv:2310.05921}, 2023.

\bibitem{lyu2024variational}
Fan Lyu, Kaile Du, Yuyang Li, Hanyu Zhao, Zhang Zhang, Guangcan Liu, and Liang
  Wang.
\newblock Variational continual test-time adaptation.
\newblock {\em arXiv preprint arXiv:2402.08182}, 2024.

\bibitem{lyu2021multi}
Fan Lyu, Shuai Wang, Wei Feng, Zihan Ye, Fuyuan Hu, and Song Wang.
\newblock Multi-domain multi-task rehearsal for lifelong learning.
\newblock In {\em Proceedings of the AAAI Conference on Artificial
  Intelligence}, volume~35, pages 8819--8827, 2021.

\bibitem{maddox2019simple}
Wesley~J Maddox, Pavel Izmailov, Timur Garipov, Dmitry~P Vetrov, and
  Andrew~Gordon Wilson.
\newblock A simple baseline for bayesian uncertainty in deep learning.
\newblock In {\em Advances in neural information processing systems},
  volume~32, 2019.

\bibitem{press2024rdumb}
Ori Press, Steffen Schneider, Matthias K{\"u}mmerer, and Matthias Bethge.
\newblock Rdumb: A simple approach that questions our progress in continual
  test-time adaptation.
\newblock {\em Advances in Neural Information Processing Systems}, 36, 2024.

\bibitem{rahimi2020post}
Amir Rahimi, Kartik Gupta, Thalaiyasingam Ajanthan, Thomas Mensink, Cristian
  Sminchisescu, and Richard Hartley.
\newblock Post-hoc calibration of neural networks.
\newblock {\em arXiv preprint arXiv:2006.12807}, 2, 2020.

\bibitem{rolnick2019experience}
David Rolnick, Arun Ahuja, Jonathan Schwarz, Timothy Lillicrap, and Gregory
  Wayne.
\newblock Experience replay for continual learning.
\newblock {\em Advances in neural information processing systems}, 32, 2019.

\bibitem{sadinle2019least}
Mauricio Sadinle, Jing Lei, and Larry Wasserman.
\newblock Least ambiguous set-valued classifiers with bounded error levels.
\newblock {\em Journal of the American Statistical Association},
  114(525):223--234, 2019.

\bibitem{shi2024controllable}
Ziqi Shi, Fan Lyu, Ye~Liu, Fanhua Shang, Fuyuan Hu, Wei Feng, Zhang Zhang, and
  Liang Wang.
\newblock Controllable continual test-time adaptation.
\newblock {\em arXiv preprint arXiv:2405.14602}, 2024.

\bibitem{sojka2023ar}
Damian S{\'o}jka, Sebastian Cygert, Bart{\l}omiej Twardowski, and Tomasz
  Trzci{\'n}ski.
\newblock Ar-tta: A simple method for real-world continual test-time
  adaptation.
\newblock In {\em Proceedings of the IEEE/CVF International Conference on
  Computer Vision}, pages 3491--3495, 2023.

\bibitem{sun2022exploring}
Qing Sun, Fan Lyu, Fanhua Shang, Wei Feng, and Liang Wan.
\newblock Exploring example influence in continual learning.
\newblock {\em Advances in Neural Information Processing Systems},
  35:27075--27086, 2022.

\bibitem{sun2020test}
Yu~Sun, Xiaolong Wang, Zhuang Liu, John Miller, Alexei Efros, and Moritz Hardt.
\newblock Test-time training with self-supervision for generalization under
  distribution shifts.
\newblock In {\em International Conference on Machine Learning}, 2020.

\bibitem{tan2024less}
Jiayao Tan, Fan Lyu, Chenggong Ni, Tingliang Feng, Fuyuan Hu, Zhang Zhang,
  Shaochuang Zhao, and Liang Wang.
\newblock Less is more: Pseudo-label filtering for continual test-time
  adaptation.
\newblock {\em arXiv preprint arXiv:2406.02609}, 2024.

\bibitem{tarvainen2017mean}
Antti Tarvainen and Harri Valpola.
\newblock Mean teachers are better role models: Weight-averaged consistency
  targets improve semi-supervised deep learning results.
\newblock In {\em Proceedings of the Advances in Neural Information Processing
  Systems}, 2017.

\bibitem{tomani2021post}
Christian Tomani, Sebastian Gruber, Muhammed~Ebrar Erdem, Daniel Cremers, and
  Florian Buettner.
\newblock Post-hoc uncertainty calibration for domain drift scenarios.
\newblock In {\em Proceedings of the IEEE/CVF Conference on Computer Vision and
  Pattern Recognition}, pages 10124--10132, 2021.

\bibitem{van2020brain}
Gido~M Van~de Ven, Hava~T Siegelmann, and Andreas~S Tolias.
\newblock Brain-inspired replay for continual learning with artificial neural
  networks.
\newblock {\em Nature communications}, 11(1):4069, 2020.

\bibitem{volkhonskiy2017inductive}
Denis Volkhonskiy, Evgeny Burnaev, Ilia Nouretdinov, Alexander Gammerman, and
  Vladimir Vovk.
\newblock Inductive conformal martingales for change-point detection.
\newblock In {\em Conformal and Probabilistic Prediction and Applications},
  pages 132--153, 2017.

\bibitem{vovk2005algorithmic}
Vladimir Vovk, Alexander Gammerman, and Glenn Shafer.
\newblock {\em Algorithmic learning in a random world}, volume~29.
\newblock Springer, 2005.

\bibitem{vovk2020computationally}
Vladimir Vovk, Ivan Petej, Ilia Nouretdinov, Valery Manokhin, and Alexander
  Gammerman.
\newblock Computationally efficient versions of conformal predictive
  distributions.
\newblock {\em Neurocomputing}, 397:292--308, 2020.

\bibitem{wang2020tent}
Dequan Wang, Evan Shelhamer, Shaoteng Liu, Bruno Olshausen, and Trevor Darrell.
\newblock Tent: Fully test-time adaptation by entropy minimization.
\newblock In {\em Proceedings of the International Conference on Learning
  Representations}, 2020.

\bibitem{wang2022continual}
Qin Wang, Olga Fink, Luc Van~Gool, and Dengxin Dai.
\newblock Continual test-time domain adaptation.
\newblock In {\em Proceedings of the Computer Vision and Pattern Recognition},
  2022.

\bibitem{xie2017aggregated}
Saining Xie, Ross Girshick, Piotr Doll{\'a}r, Zhuowen Tu, and Kaiming He.
\newblock Aggregated residual transformations for deep neural networks.
\newblock In {\em Proceedings of the Computer Vision and Pattern Recognition},
  2017.

\bibitem{yang2023exploring}
Xu~Yang, Yanan Gu, Kun Wei, and Cheng Deng.
\newblock Exploring safety supervision for continual test-time domain
  adaptation.
\newblock In {\em Proceedings of the International Joint Conference on
  Artificial Intelligence}, 2023.

\bibitem{yilmaz2022test}
Fatih~Furkan Yilmaz and Reinhard Heckel.
\newblock Test-time recalibration of conformal predictors under distribution
  shift based on unlabeled examples.
\newblock {\em arXiv preprint arXiv:2210.04166}, 2022.

\bibitem{zaffran2022adaptive}
Margaux Zaffran, Olivier F{\'e}ron, Yannig Goude, Julie Josse, and Aymeric
  Dieuleveut.
\newblock Adaptive conformal predictions for time series.
\newblock In {\em International Conference on Machine Learning}, pages
  25834--25866, 2022.

\bibitem{zagoruyko2016wide}
Sergey Zagoruyko and Nikos Komodakis.
\newblock Wide residual networks.
\newblock In {\em Procedings of the British Machine Vision Conference}, 2016.

\bibitem{zou2024coverage}
Xin Zou and Weiwei Liu.
\newblock Coverage-guaranteed prediction sets for out-of-distribution data.
\newblock In {\em Proceedings of the AAAI Conference on Artificial
  Intelligence}, volume~38, pages 17263--17270, 2024.

\end{thebibliography}

% \appendix

% \section{Appendix 1}

\end{document}